\documentclass[anonymous=false, sigconf=true, review=false, natbib=true]{acmart}

\AtBeginDocument{%
  \providecommand\BibTeX{{%
    \normalfont B\kern-0.5em{\scshape i\kern-0.25em b}\kern-0.8em\TeX}}}

\copyrightyear{2023}
\acmYear{2023}
\setcopyright{acmlicensed}\acmConference[SIGIR '23]{Proceedings of the 46th International ACM SIGIR Conference on Research and Development in Information Retrieval}{July 23--27, 2023}{Taipei, Taiwan}
\acmBooktitle{Proceedings of the 46th International ACM SIGIR Conference on Research and Development in Information Retrieval (SIGIR '23), July 23--27, 2023, Taipei, Taiwan}
\acmPrice{15.00}
\acmDOI{10.1145/3539618.3591918}
\acmISBN{978-1-4503-9408-6/23/07}

\usepackage{array, multirow}
\usepackage{enumitem}
\usepackage{amsfonts} 
\usepackage[capitalise]{cleveref}
\usepackage{caption}
\usepackage{subcaption}

\newcolumntype{L}{>{\centering\arraybackslash}m{3cm}}

\begin{document}

\newcommand{\lacaml}{LA$_{\textsc{caml}}$}
\newcommand{\lalaat}{LA$_{\textsc{laat}}$}

\title[Automated Medical Coding on MIMIC-III and MIMIC-IV]{Automated Medical Coding on MIMIC-III and MIMIC-IV: A Critical Review and Replicability Study}

\author{Joakim Edin}
\orcid{0000-0003-1005-8276}
\affiliation{%
  \institution{University of Copenhagen}
  \country{}
  }
\affiliation{%
  \institution{Corti}
  \country{}
}
\email{je@corti.ai}

\author{Alexander Junge}
\orcid{0000-0002-2410-9671}
\affiliation{%
  \institution{Corti}
  \country{}
}
\email{aju@corti.ai}

\author{Jakob D. Havtorn}
\orcid{0000-0002-4849-0817}
\affiliation{%
  \institution{Technical University of Denmark}
  \country{}
}
\affiliation{%
  \institution{Corti}
  \country{}
}
\email{jdh@corti.ai}

\author{Lasse Borgholt}
\orcid{0000-0002-3562-8442}
\affiliation{%
  \institution{Aalborg University}
  \country{}
}
\affiliation{%
  \institution{Corti}
  \country{}
}
\email{lb@corti.ai}

\author{Maria Maistro}
\affiliation{%
  \institution{University of Copenhagen}
  \country{}
}
\email{mm@di.ku.dk}

\author{Tuukka Ruotsalo}
\affiliation{%
  \institution{University of Copenhagen}
  \country{}
}
\affiliation{%
  \institution{University of Helsinki}
  \country{}
}
\email{tr@di.ku.dk}

\author{Lars Maal{\o}e}
\affiliation{%
  \institution{Technical University of Denmark}
  \country{}
}
\affiliation{%
  \institution{Corti}
  \country{}
}
\email{lm@corti.ai}

\renewcommand{\shortauthors}{Edin et al.}

\begin{abstract}
Medical coding is the task of assigning medical codes to clinical free-text documentation.
Healthcare professionals manually assign such codes to track patient diagnoses and treatments. Automated medical coding can considerably alleviate this administrative burden. In this paper, we reproduce, compare, and analyze state-of-the-art automated medical coding machine learning models. We show that several models underperform due to weak configurations, poorly sampled train-test splits, and insufficient evaluation. In previous work, the macro F1 score has been calculated sub-optimally, and our correction doubles it. We contribute a revised model comparison using stratified sampling and identical experimental setups, including hyperparameters and decision boundary tuning. We analyze prediction errors to validate and falsify assumptions of previous works. The analysis confirms that all models struggle with rare codes, while long documents only have a negligible impact. Finally, we present the first comprehensive results on the newly released MIMIC-IV dataset using the reproduced models. We release our code, model parameters, and new MIMIC-III and MIMIC-IV training and evaluation pipelines to accommodate fair future comparisons.\footnote{\label{footnote:source_code}\url{https://github.com/JoakimEdin/medical-coding-reproducibility}}

\end{abstract}

\ccsdesc[500]{Information systems~Information retrieval}

\keywords{Automated Medical Coding; Reproducibility; MIMIC}



\maketitle

\begin{table*}[t!]
    \centering
    \caption{Comparison of the previously defined MIMIC-III splits \textit{full} and \textit{50} \cite{mullenbachExplainablePredictionMedical2018} and our proposed MIMIC-III \textit{clean} split along with similarly defined splits for MIMIC-IV \textit{ICD-9} and \textit{ICD-10} after pre-processing.}
    \label{tab:subsets}
    \begin{tabular}{lccccc}
        \toprule
        & \multicolumn{2}{c}{\textit{Previous work}} & \multicolumn{3}{c}{\textit{Our work}} \\
        \cmidrule(lr){2-3}\cmidrule(lr){4-6}
        & {MIMIC-III \textit{full}} & {MIMIC-III 50}  & {MIMIC-III \textit{clean}}  & {MIMIC-IV \textit{ICD-9}}  & {MIMIC-IV \textit{ICD-10}}\\
        \midrule
        Number of documents  & 52,723 & 11,368 &  52,712 & 209,326 & 122,279 \\
        Number of patients & 41,126 & 10,356 & 41,118 & 97,709 & 65,659 \\
        Number of unique codes  & 8,929 & 50  & 3,681 & 6,150 & 7,942\\
        Codes pr. instance: Median (IQR)    & 14 (10-20) & 5 (3-8) & 14 (10-20) & 12 (8-17) & 14 (9-20)\\
        Words pr. document: Median (IQR) & 1,375 (965-1,900) & 1,478 (1,065-1,992) & 1,311 (917-1,822) & 1,320 (997-1,715) & 1,492 (1,147-1,931)\\
        Documents: Train/val/test [\%] & 90.5/3.1/6.4 & 71.0/13.8/15.2 & 72.9/10.6/16.6  & 73.8/10.5/15.7& 72.9/10.9/16.2  \\
        Missing codes: Train/val/test [\%] & 2.7/66.4/54.3 & 0.0/0.0/0.0 & 0.0/0.1/0.0 & 0.0/0.5/0.2 & 0.0/0.5/0.1\\
        \bottomrule
    \end{tabular}
\end{table*}

\section{Introduction}

Medical coding is the task of assigning diagnosis and procedure codes to free-text medical documentation \cite{dongAutomatedClinicalCoding2022,tengReviewDeepNeural2022}.
These codes ensure that patients receive the correct level of care and that healthcare providers are accurately compensated for their services. However, this is a costly manual process prone to error \cite{tsengAdministrativeCostsAssociated2018, omalleyMeasuringDiagnosesICD2005, burnsSystematicReviewDischarge2012}.

The goal of automated medical coding (AMC) is to predict a set of codes or provide a list of codes ranked by relevance for a medical document. Numerous machine learning models have been developed for AMC \cite{jiUnifiedReviewDeep2022, tengReviewDeepNeural2022, stanfillSystematicLiteratureReview2010}. These models are trained on datasets of medical documents, typically discharge summaries, each labeled with a set of medical codes.
While some models treat AMC as an ad-hoc information retrieval problem \cite{rizzoICDCodeRetrieval2015,parkInformationRetrievalApproach2019}, it is more commonly posed as a multi-label classification problem \cite{tengReviewDeepNeural2022, jiUnifiedReviewDeep2022}.

While most research in AMC has been conducted on the third version of the Medical Information Mart for Intensive Care dataset (MIMIC-III) \cite{tengReviewDeepNeural2022, venkateshAutomatingOverburdenedClinical2023}, it remains challenging to compare the results of different models. Performance improvements are commonly attributed to model design, but differences in experimental setups make these claims hard to validate.
In addition, long documents, rare codes, and lack of training data are often cited as core research challenges \cite{baoMedicalCodePrediction2021,dongAutomatedClinicalCoding2022,dongExplainableAutomatedCoding2021,feuchtDescriptionbasedLabelAttention2021,gaoLimitationsTransformersClinical2021,huangPLMICDAutomaticICD2022,jiDoesMagicBERT2021,jiUnifiedReviewDeep2022,kavuluruEmpiricalEvaluationSupervised2015,kimReadAttendCode2021,liICDCodingClinical2020,liuEffectiveConvolutionalAttention2021,michalopoulosICDBigBirdContextualEmbedding2022,moonsComparisonDeepLearning2020,pascualBERTbasedAutomaticICD2021,tengReviewDeepNeural2022,tengExplainablePredictionMedical2020,xieEHRCodingMultiscale2019,yangKnowledgeInjectedPrompt2022,zhangBERTXMLLargeScale2020,zhouAutomaticICDCoding2021, vuLabelAttentionModel2020, venkateshAutomatingOverburdenedClinical2023}. However, except for a few studies demonstrating performance drops on rare codes, the number of studies containing in-depth error analyses is limited \cite{baoMedicalCodePrediction2021,dongExplainableAutomatedCoding2021,jiDoesMagicBERT2021}.

We address the above challenges. Our major contributions are:
\begin{itemize}[leftmargin=5mm]
    \item[1)] We reproduce the performance of state-of-the-art models on MIMIC-III. We find that evaluation methods are flawed and propose corrections that double the macro F1 scores.
    \item[2)] We find the original split of MIMIC-III to introduce strong biases in results due to missing classes in the test set. We create a new split with full class representation using stratified sampling.
    \item [3)] We perform a revised model comparison on MIMIC-III \textit{clean} using the same training, evaluation, and experimental setup for all models. We find that models previously reported as low-performing improve considerably, demonstrating the importance of hyperparameters and decision boundary tuning.
    \item[4)] We report the first results of current state-of-the-art models on the newly released MIMIC-IV dataset \cite{johnsonMIMICIVFreelyAccessible2023, goldbergerPhysioBankPhysioToolkitPhysioNet2000}. We find that previous conclusions generalize to the new dataset.
    \item[5)] Through error analysis, we provide empirical evidence for multiple model weaknesses. Most importantly, we find that rare codes harm performance, while, in contrast to previous claims, long documents only have a negligible performance impact.
  
\end{itemize}
We release our source code and new splits for MIMIC-III and IV$^\text{1}$
and hope these contributions will aid future research in AMC.

    


\section{Previous work}

In the following, we review datasets, model architectures, training, and evaluation of the models we compare in this study. Our criteria for selecting these models are presented in \cref{sec: inclusion and exclusion criteria}.

\subsection{Datasets}
\label{sec:dataset}
The International Classification of Diseases (ICD) is the most popular medical coding system worldwide \cite{tengReviewDeepNeural2022}. It follows a tree-like hierarchical structure, also known as a medical ontology, to ensure the functional and structural integrity of the classification. Chapters are the highest level in the hierarchy, followed by categories, sub-categories, and codes. The World Health Organization (WHO) revises ICD periodically. Each revision introduces new codes. For instance, ICD-9 contains 18,000 codes, while ICD-10 contains 142,000.\footnote{\url{https://www.cdc.gov/nchs/icd/icd10cm_pcs_background.htm}} MIMIC-II and MIMIC-III are the most widely used open-access datasets for research on ICD coding and are provided by the Beth Israel Deaconess Medical Center \cite{tengReviewDeepNeural2022, johnsonMIMICIIIFreelyAccessible2016, leeOpenaccessMIMICIIDatabase2011}. 
 
MIMIC-III contains medical documents annotated with ICD-9 codes collected between 2001 and 2012 \cite{johnsonMIMICIIIFreelyAccessible2016}. Usually, discharge summaries---free-text notes on patient and hospitalization history---are the only documents used for AMC \cite{tengReviewDeepNeural2022}. MIMIC-III \textit{full} and \textit{50} are commonly used splits. MIMIC-III \textit{full} contains all ICD-9 codes, while \textit{50} only contains the top 50 most frequent codes \cite{mullenbachExplainablePredictionMedical2018, shiAutomatedICDCoding2018}. 

\begin{figure}[b]
    \centering
    \includegraphics[width=0.9\linewidth]{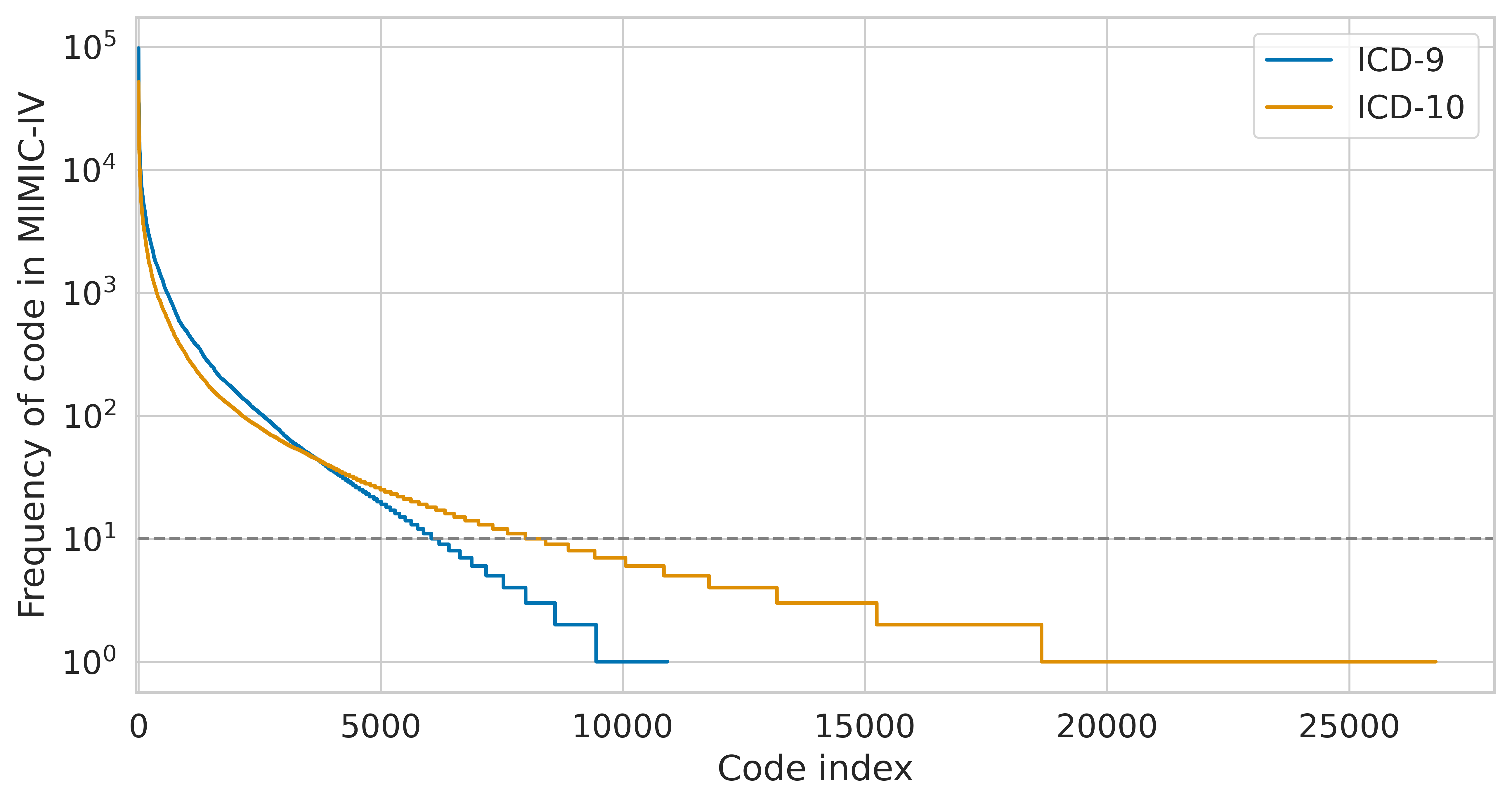}
    \caption{The frequency of ICD-9 and ICD-10 codes in MIMIC-IV before pre-processing. As discussed in \cref{sec: splits}, we removed codes with fewer than ten instances (dashed line).}
    \label{fig:mimiciv_code_dist}
\end{figure}

MIMIC-IV was released on January 6th, 2023, and has not previously been used for AMC. It contains data for patients admitted to the Beth Israel Deaconess Medical Center emergency department or ICU between 2008-2019  annotated with either ICD-9 or ICD-10 codes \cite{johnsonMIMICIVFreelyAccessible2023}. The empirical frequencies of codes of each ICD version in MIMIC-IV are shown in \cref{fig:mimiciv_code_dist}.  
Statistics for the MIMIC-III \textit{50}, \textit{full}, and MIMIC-IV datasets are listed in \cref{tab:subsets}.

\subsection{Model architectures}

Most recent state-of-the-art AMC models use an encoder-decoder architecture. The encoder takes a sequence of tokens $T \in \mathbb{Z}^{n}$ as input and outputs a sequence of hidden representations $H \in \mathbb{R}^{d_h \times n}$, where $n$ is the number of tokens in a sequence, and $d_h$ is the hidden dimension. The decoder takes $H$ as input and outputs the code probability distributions. For the task of ranking, codes are sorted by decreasing probability. For classification, code probabilities larger than a set decision boundary are predicted. 


\subsubsection{Encoders:}
The encoder usually consists of pre-trained non-contextualized word embeddings (e.g., Word2Vec) and a neural network for encoding context. More recently, pre-trained masked language models (e.g., BERT) have gained popularity~\cite{tengReviewDeepNeural2022}. The MIMIC-III training set or PubMed articles are commonly used for pre-training.

\subsubsection{Decoders:}
The most common decoder architectures can be grouped into three primary types. 
The simplest decoder is a pooling layer (e.g., max pooling) followed by a feed-forward neural network. More recently, label-wise attention (LA) \cite{mullenbachExplainablePredictionMedical2018} has replaced pooling \cite{vuLabelAttentionModel2020, liuEffectiveConvolutionalAttention2021, huangPLMICDAutomaticICD2022, liICDCodingClinical2020}. LA transforms a sequence of hidden representations $H$ into label-specific representations $V \in \mathbb{R}^{d_h \times L}$, where $L$ is the number of unique medical codes in the dataset.
It is computed as
\begin{equation}
    A = \text{softmax}(WH) \, , \quad
    V = HA^{\top} \enspace ,
\end{equation}
where the softmax normalizes each column of $WH$, $W \in \mathbb{R}^{L \times d_h}$ is an embedding matrix that learns label-specific queries, and $A \in \mathbb{R}^{L \times n}$ is the attention matrix. 
Then, $V$ is used to compute class-wise probabilities via a feedforward neural network. 
As LA was first used in the \textit{convolutional attention for multi-label classification} (CAML) model~\cite{mullenbachExplainablePredictionMedical2018}, we refer to this method as \lacaml.

An updated label-wise attention module was introduced in the \textit{label attention model} (LAAT) \cite{vuLabelAttentionModel2020}. We refer to this attention module as \lalaat. In \lalaat, the label-specific attention is computed similarly to  \lacaml~ as $A = \text{softmax}(UZ)$, where $U \in \mathbb{R}^{L \times d_p}$ is a learnable embedding matrix, but with 
$Z = \text{tanh}(PH)$ 
where $P \in \mathbb{R}^{d_p \times d_h}$ is a learnable matrix, $Z \in \mathbb{R}^{d_p \times n}$ and $d_p$ is a hyperparameter.

\begin{table}[t]
    \centering
    \caption{An overview of the compared models.}
    \label{tab:model_facts}
    \begin{tabular}{lcccc}
        \toprule
        Model   & Encoder & Decoder  & Param \\
        \midrule
        Bi-GRU \cite{mullenbachExplainablePredictionMedical2018}   &  Word2Vec, Bi-GRU & Max-pool  &  9.9M\\
        CNN \cite{mullenbachExplainablePredictionMedical2018}   & Word2Vec, CNN & Max-pool &  10.3M\\
        CAML \cite{mullenbachExplainablePredictionMedical2018}   & Word2Vec, CNN  & \lacaml &  6.1M\\
        MultiResCNN \cite{liICDCodingClinical2020} &  Word2Vec, ResNet & \lacaml &  11.9M\\
        LAAT \cite{vuLabelAttentionModel2020} & Word2Vec, Bi-LSTM  & \lalaat  &  21.9M\\
        PLM-ICD \cite{huangPLMICDAutomaticICD2022} &  BERT & \lalaat &  138.8M\\
        \bottomrule
    \end{tabular}
\end{table}

\subsection{Training and evaluation methods}\label{sec: training and evaluation methods}
\citet{mullenbachExplainablePredictionMedical2018} released code for pre-processing the discharge summaries, generating the train-test split, and evaluating model performance on MIMIC-III which many subsequent papers have used \cite{vuLabelAttentionModel2020, huangPLMICDAutomaticICD2022, liICDCodingClinical2020, baoMedicalCodePrediction2021, yuanCodeSynonymsMatter2022, kimReadAttendCode2021}.
Pre-processing consisted of lowercasing all text and removing words that only contain out-of-alphabet characters. Predicting procedures and diagnosis codes were treated as a single task.
The dataset was split into training, validation, and test sets using random sampling, ensuring that no patient occurred in both the training and test set. The (non-stratified) random sampling lead to 54\% of the ICD codes in MIMIC-III \textit{full} not being sampled in the test set. This complicates the interpretation of results since these codes only contribute true negatives or false positives. 
Models are evaluated using the micro and macro average of the area under the curve of the receiver operating characteristics (AUC-ROC), F1 score, and Precision@k. 

While most papers use the pre-processing, train-test split, and evaluation metrics described above, they differ in several aspects of training. This may lead to performance differences unrelated to modeling choices which are undesirable when seeking to compare models. 
For instance, due to varying memory constraints of different models, documents are usually truncated to some maximum length. In the literature, this maximum varies from 2,500 to 4,000 words \cite{mullenbachExplainablePredictionMedical2018, vuLabelAttentionModel2020, huangPLMICDAutomaticICD2022}.
Furthermore, not all papers tune the prediction decision boundary but simply set it to 0.5,
 hyperparameter search ranges and sampling methods  vary between works, and learning rate schedulers are only used in LAAT and PLM-ICD\cite{mullenbachExplainablePredictionMedical2018, liICDCodingClinical2020}. In LAAT, the learning rate was decreased by 90\% when the F1 score had not increased for five epochs. PLM-ICD used a schedule with linear warmup followed by linear decay. 


All models except for PLM-ICD use Word2Vec embeddings pre-trained on the MIMIC-III training set. PLM-ICD uses a BERT encoder pre-trained on PubMed to encode the text in chunks of 128 tokens, and these contextualized embeddings are fed to a \lalaat~ layer. 

Finally, all models compute independent code probabilities using sigmoid activation functions and optimize the binary cross entropy loss function during training.
\cref{tab:model_facts} presents the selected models.

\section{Methods}
\subsection{Selection criteria}\label{sec: inclusion and exclusion criteria}

In this study, we included both models trained from scratch and models with components pre-trained on external corpora. 
We excluded models that use multi-modal inputs, such as medical code descriptions \cite{kimReadAttendCode2021, mullenbachExplainablePredictionMedical2018, vuLabelAttentionModel2020, caoHyperCoreHyperbolicCograph2020, baoMedicalCodePrediction2021}, code synonyms \cite{yuanCodeSynonymsMatter2022}, code hierarchies \cite{caoHyperCoreHyperbolicCograph2020, xieEHRCodingMultiscale2019}, or associated Wikipedia articles \cite{baiImprovingMedicalCode2019}, because they introduced additional complexity without providing evidence for significant performance improvements \cite{mullenbachExplainablePredictionMedical2018, vuLabelAttentionModel2020, tengReviewDeepNeural2022}. We excluded works without publically available source code as the experiment descriptions often lacked important implementation details.

\subsection{Evaluation metrics}
\label{sec:metrics}
Similar to previous work, we evaluated models using AUC-ROC, F1 score, and precision@$k$. Additionally, we introduced exact match ratio (EMR), R-precision, and mean average precision (MAP). The EMR is the percentage of instances where all codes were predicted correctly. This allowed us to measure how many documents were predicted perfectly, which is important for \textit{fully automated} medical coding.
Where precision@$k$ is computed based on the top-$k$ codes (i.e., $k$ is fixed), R-precision considers a number of codes equal to the true number of relevant codes. Thus, R-precision is useful when the number of relevant codes varies considerably between documents, which is the case for the MIMIC datasets. Finally, in contrast to all other metrics, MAP considers the exact rank of all relevant codes in a document.


\begin{table*}[t]
    \centering
    \caption{Hyperparameters, maximum document lengths, and decision boundary tuning strategies used in the original works compared to the optimal settings found in this paper (marked with *). LR is the learning rate scheduler. ``Length" is the maximum number of words a document can contain before being truncated. $\dagger$ applies to models using word-piece tokenization. These models were filtered on the number of sub-words instead of full words. ``DB tune" is whether the optimal decision boundary was found using the validation set. If a paper did not tune the decision boundary, it was set to 0.5.}
    \label{tab:hyperparameters}
    \begin{tabular}{lccccccccc}
        \toprule
        & \multicolumn{7}{c}{Hyperparameters} & & \\
        \cmidrule(lr){2-8}
        Model & Batch Size & Weight Decay & Learning Rate & Dropout & LR Scheduler & Optimizer & Epochs & Length & DB tune \\
        \midrule
        Bi-GRU  & 16 &  0.0 & 0.003 &  0.2   & no  & Adam &  100 & 2500 & no\\
        Bi-GRU$^*$  & 8 &  0.0001 & 0.001 &  0   & yes  & AdamW &  20 & 4000 & yes\\
        \hline
        CNN  & 16 &  0.0 & 0.003  &  0.2 & no  & Adam &  100 & 2500 & no\\
        CNN$^*$    & 8 &  0.00001 & 0.001  &  0 & yes  & AdamW &  20 & 4000 & yes \\
        \hline
        CAML  & 16 &  0.0 & 0.0001  &  0.2  & no  & Adam  &  200 & 2500 & no\\
        CAML$^*$    & 8 &  0.001 & 0.005  &  0.2  & yes  & AdamW  &  20 & 4000 & yes\\
        \hline
        MultiResCNN & 16 &  0.0 & 0.0001  &  0.2  & no  & Adam &  200 & 2500 & no\\
        MultiResCNN$^*$  & 16 &  0.0001 & 0.0005  &  0.2  & yes  & AdamW &  20 & 4000 & yes\\
        \hline
        LAAT   & 8 &  0.0 & 0.0001  &  0.3  & yes  & AdamW &  50 & 4000 & no\\
        LAAT$^*$    & 8 &  0.001 & 0.001  &  0.2  & yes  & AdamW &  20 & 4000 & yes\\
        \hline
        PLM-ICD  & 8 &  0.0 & 0.00005  &  0.2  & yes  & AdamW &  20 & 3072$^\dagger$ & yes \\
        PLM-ICD$^*$   & 16 &  0.0 & 0.00005  &  0.2  & yes  & AdamW &  20 & 4000 & yes \\
        \bottomrule
    \end{tabular}
\end{table*}

Previous works calculated the macro F1 score as the harmonic mean of the macro precision and macro recall \cite{mullenbachExplainablePredictionMedical2018, liICDCodingClinical2020, vuLabelAttentionModel2020, huangPLMICDAutomaticICD2022}. \citet{opitzMacroF1Macro2021} analyze macro F1 formulas common in multi-class and multi-label classification. They demonstrate that the above formulation is sub-optimal, as it rewards heavily biased classifiers in unbalanced datasets. Therefore, as recommended by the authors, we calculated the macro F1 score as the arithmetic mean of the F1 score for each class.
As seen in \cref{tab:subsets}, 54\% of codes in MIMIC-III \textit{full} are missing in the test set. Previous works set the F1 score of all the missing codes in the test set to 0, resulting in a misleadingly low macro F1 score. Because 54\% of the codes are missing, the maximum possible macro F1 score is 46\%. 
We ignored all codes not in the test set for our reproduction, essentially trading bias for variance. 
For our revised comparison, we resolved the issue by instead sampling new splits that reduce missing codes to a negligible fraction (see \cref{sec: splits}) and ignoring the few that were still missing.

\subsection{Definition of splits}
\label{sec: splits}


We define three new splits: MIMIC-III \textit{clean}, MIMIC-IV \textit{ICD-9}, and \textit{ICD-10}.
As described in \cref{sec:metrics}, 54\% of the codes in MIMIC-III \textit{full} are absent from the test set, which introduces significant bias in the model evaluation metrics. Therefore, we created a new MIMIC-III split to ensure that most codes are present in both the training and test set. 
Specifically, we removed codes with fewer than ten occurrences, doubled the test set size, and sampled the documents using multi-label stratified sampling \cite{sechidisStratificationMultilabelData2011}. 
We ensured that no patient occurred in both the training and test set, preprocessed the text, and considered procedures and diagnosis codes as a single task as done by \citet{mullenbachExplainablePredictionMedical2018}.
We based our new split on the v1.4 version of the dataset and refer to it as MIMIC-III \textit{clean}.
Using the same method, we created two splits for MIMIC-IV v2.2: one containing all documents labeled with ICD-9 codes and one with ICD-10 codes.

\subsection{Reproducibility experiments}

We ran reproducibility experiments with all models to evaluate whether the results in the original works could be reproduced and to validate our reimplementations. We ran these experiments on MIMIC-III \textit{full}, and \textit{50} as in the original works \cite{mullenbachExplainablePredictionMedical2018,liICDCodingClinical2020,vuLabelAttentionModel2020,huangPLMICDAutomaticICD2022}. We used the hyperparameters reported in each paper (see \cref{tab:hyperparameters}) and report both the original and the revised macro F1 scores discussed in \cref{sec:metrics}.

\subsection{Revised comparison}

To address the issues associated with comparing results reported by previous works described in \cref{sec: training and evaluation methods,sec:metrics}, we perform a revised model comparison. We run experiments on the new MIMIC-III \textit{clean}, MIMIC-IV \textit{ICD-9}, and \textit{ICD-10}.
All models were trained for 20 epochs using a learning rate schedule with linear warmup for the first 2K updates followed by linear decay \cite{huangPLMICDAutomaticICD2022}. We found this schedule to speed up the training convergence of all the models.
Whereas original works use Adam or AdamW, we used AdamW for all experiments as it corrects the weight decay implementation of Adam \cite{kingmaAdamMethodStochastic2017, loshchilovDecoupledWeightDecay2022}. For each model, we tuned the decision boundary to maximize the micro F1 score on the validation set. We used randomized sampling to find optimal settings for dropout, weight decay, learning rate, and batch size. The hyperparameter search was performed on MIMIC-III \textit{clean}, and the MIMIC-IV splits. We found that the best setting for each model generalized across datasets. Using this setting, we ran each model ten times with different seeds on each dataset. All documents were truncated to a maximum of 4000 words. The hyperparameters, maximum document lengths, and decision boundary tuning strategy are summarised in \cref{tab:hyperparameters}. 


We performed an ablation study to analyze which changes had the largest impact on performance. 
Specifically, we evaluated the effect of truncation, hyperparameter search, and decision boundary tuning. We modified one of these at a time: We ran one experiment where documents were truncated to a maximum length of 2,500 words, a second experiment where the models were trained with the hyperparameters, number of epochs, and learning rate schedule used in the original works, and a third experiment where the decision boundary was set to 0.5 instead of tuned.

\subsection{Error analysis}
To validate and falsify the commonly cited challenges of AMC, which include a lack of training data, long documents, and rare codes, we performed an error analysis. In addition to analyzing rare codes, we contribute an in-depth code analysis aiming to identify the attributes that make certain codes challenging to predict.

\begin{table*}[t]
\centering
\caption{Reproduced test set results compared with those from the original works. Our reproduced results are indicated with *. The results were reproduced on MIMIC-III v1.4 with the preprocessing pipeline and splits of \citet{mullenbachExplainablePredictionMedical2018}. Each model was reproduced using the hyperparameters presented in the respective paper. We use both macro F1 formulas: Macro$^\dagger$ refers to the method used in the original work, while Macro refers to the corrected version used in this paper.}
\label{tab:reproduced_results}
\begin{tabular}{l  cc  ccc  cc  cc  ccc  c}
    \toprule
     & \multicolumn{7}{c}{MIMIC-III \textit{full}} & \multicolumn{6}{c}{MIMIC-III \textit{50}} \\
     \cmidrule(lr){2-8}\cmidrule(lr){9-14}
     & \multicolumn{2}{c}{AUC-ROC} & \multicolumn{3}{c}{F1} & \multicolumn{2}{c}{Precision@$k$}  & \multicolumn{2}{c}{AUC-ROC} & \multicolumn{3}{c}{F1} & Precision@$k$  \\
     & Micro & Macro & Micro & Macro$^\dagger$ & Macro & 8 & 15 & Micro & Macro & Micro & Macro$^\dagger$ & Macro & 5 \\
    \midrule
    CNN &  96.9 & 80.6 &  41.9& 4.2 & -& 58.1 &  48.8  & 90.7 & 87.6 & 62.5 & 57.6 & - & 62.0  \\
    CNN$^*$ & 97.3 & 83.1 & 41.5 & 3.4 & 6.7 & 61.9 & 47.2 & 91.9 & 89.2 & 64.9 & 58.8 & 58.0 &  62.6 \\
    \hline
    Bi-GRU & 97.1 & 82.2 & 41.7 & 3.8 & - & 58.5 & 44.5  &  89.2 & 82.8 & 54.9 &  48.4 & - &  59.1  \\
    Bi-GRU$^*$  & 98.0 & 87.1 & 42.6 & 3.6 & 7.0 & 65.0 & 49.8 & 89.3 & 85.2 & 56.1 & 46.2 & 43.1 & 57.9  \\
    \hline
    CAML  &  98.6 &  89.5 &  53.9 & 8.8 & - &  70.9 &  56.1  & 90.9 & 87.5 &  61.4 &  53.2 & - &   60.9  \\
    CAML$^*$ & 98.4 & 88.4 & 49.5 & 5.6 & 11.3 & 69.9 & 54.9 & 91.1 & 87.5 & 60.6 & 52.4 & 51.0 & 61.1  \\
    \hline
    MultiResCNN  &  98.6 &  91.0 & 55.2 & 8.6 & - & 73.4 & 58.4  &  93.8 &  89.9 &  67.0 &  60.6 & - &  64.1 \\
    MultiResCNN$^*$  & 98.6 & 90.8 & 56.5 & 9.2 & 18.5 & 73.4 & 58.4 & 92.4 & 89.7 & 67.3 & 62.2 & 61.1 & 63.4  \\
    \hline
    LAAT  &  98.8 &  91.9 &  57.5 & 9.9 & - &  74.5 &  59.1  &  94.6 &  92.5 &  71.5 &  66.6 & - &  67.5 \\
    LAAT$^*$ & 98.6 & 89.5 & 56.1 & 8.2 & 16.2 & 73.9 & 58.7 & 92.8 & 90.5 & 66.8 & 60.8 & 59.2 & 64.0  \\
    \hline
    PLM-ICD  &  98.9 & 92.6 &  59.8 & 10.4 & - &  77.1 &  61.3  & - & - & - & - & - & -  \\
    PLM-ICD$^*$ & 98.8 & 92.3 & 58.9 & 11.1 & 22.8 & 75.7 & 60.5 & 93.8 & 91.7 & 70.5 & 66.3 & 65.4 & 65.7 \\
    \bottomrule
\end{tabular}
\end{table*}

\subsubsection{Amount of training data:}
Multiple studies attribute poor performance to data sparsity of MIMIC-III, which contains only fifty thousand examples \cite{kavuluruEmpiricalEvaluationSupervised2015,tengExplainablePredictionMedical2020,yanMedicalCodingClassification2010,yangKnowledgeInjectedPrompt2022}. MIMIC-IV \textit{ICD-9} contains four times as many examples, which allows analyzing the effect of training set size. We train each model on 25k, 50k, 75k, 100k, and 125k examples and report micro and macro F1 on the fixed test set. The training subsets were sampled from the training set using multi-label stratified sampling to ensure the same code distributions \cite{sechidisStratificationMultilabelData2011}.


\subsubsection{Document length:}
We analyzed whether model performance correlates with document length on MIMIC-IV \textit{ICD-9}. Specifically, we calculated the Pearson and Spearman correlation between the number of words in the documents and the micro F1 score for all models. For each model, we used the best seed from the revised comparison.

\subsubsection{Code analysis:}
To analyze the performance impact of rare codes, we first calculated the Pearson and Spearman correlation between model performance on each code and the corresponding code frequency in the training data. We calculated these correlations for all splits. To identify attributes of challenging codes, we analyzed model performance on the chapter level of the ICD-10 classification system. Using high-level chapters instead of codes allows us to group examples into categories, which we use as a starting point for further analysis. We limit the scope of the analysis to diagnosis codes. We focused on ICD-10 because it is the classification system currently in use at most hospitals.

\section{Results}
\subsection{Reproduced results}

In \cref{tab:reproduced_results}, we report the reproduced results on MIMIC-III \textit{full} and \textit{50} using hyperparameters as reported in the original papers. We list the original and corrected macro F1 score described in \cref{sec:metrics}. In most cases, our corrections doubled the macro F1 scores on MIMIC-III \textit{full}. The differences were smaller on MIMIC-III \textit{50} because all included codes are in the test set.

\begin{table*}[t]
    \centering
    \caption{Results on the MIMIC-III \textit{clean}, MIMIC-IV \textit{ICD-9} and MIMIC-IV  \textit{ICD-10} test sets presented as percentages. Micro F1 scores rank the table in ascending order. Each model was trained ten times with different seeds. We performed a McNemar's test with Bonferroni correction and found that all the models are significantly different ($p < 0.001$).}
    \label{tab:fair_results}
    \begin{tabular}{llccccccccc}
        \toprule
        & & \multicolumn{5}{c}{Classification} & \multicolumn{4}{c}{Ranking} \\ 
        \cmidrule(lr){3-7}\cmidrule(lr){8-11}
         & & \multicolumn{2}{c}{AUC-ROC} & \multicolumn{2}{c}{F1} & EMR & \multicolumn{2}{c}{Precision@k} & R-precision & MAP \\
         & & Micro & Macro & Micro & Macro &  & 8 & 15 &  &  \\
        \midrule
        \multirow{6}{*}{\shortstack[c]{MIMIC-III \\ clean}} & CNN & 97.1$\pm$0.0 & 88.1$\pm$0.2 & 48.0$\pm$0.3 & 9.9$\pm$0.4 & 0.1$\pm$0.0 & 61.6$\pm$0.2 & 46.6$\pm$0.1 & 49.1$\pm$0.2 & 50.6$\pm$0.2 \\
        & Bi-GRU  & 97.8$\pm$0.1 & 91.1$\pm$0.2 & 49.7$\pm$0.4 & 12.2$\pm$0.2 & 0.1$\pm$0.0 & 62.8$\pm$0.4 & 47.6$\pm$0.4 & 50.1$\pm$0.4 & 52.1$\pm$0.4 \\
        & CAML & 98.2$\pm$0.0 & 91.4$\pm$0.2 & 55.4$\pm$0.1 & 20.4$\pm$0.3 & 0.1$\pm$0.0 & 67.7$\pm$0.2 & 52.8$\pm$0.1 & 55.8$\pm$0.1 & 58.9$\pm$0.2 \\
        & MultiResCNN & 98.5$\pm$0.0 & 93.1$\pm$0.3 & 56.4$\pm$0.2 & 22.9$\pm$0.6 & 0.1$\pm$0.0 & 68.5$\pm$0.2 & 53.5$\pm$0.1 & 56.7$\pm$0.2 & 59.9$\pm$0.3 \\
        & LAAT & 98.6$\pm$0.1 & 94.0$\pm$0.3 & 57.8$\pm$0.2 & 22.6$\pm$0.6 & 0.2$\pm$0.1 & 70.1$\pm$0.2 & 54.8$\pm$0.2 & 58.0$\pm$0.2 & 61.7$\pm$0.3 \\
        & PLM-ICD & \bfseries 98.9$\pm$0.0 & \bfseries 95.9$\pm$0.1 & \bfseries 59.6$\pm$0.2 & \bfseries 26.6$\pm$0.8 & \bfseries 0.4$\pm$0.0 & \bfseries 72.1$\pm$0.2 & \bfseries 56.5$\pm$0.1 & \bfseries 60.1$\pm$0.1 & \bfseries 64.6$\pm$0.2 \\
        \hline
        \multirow{6}{*}{\shortstack[c]{MIMIC-IV \\ ICD-9}} & CNN  & 98.1$\pm$0.1 & 89.4$\pm$0.5 & 52.4$\pm$0.1 & 12.6$\pm$0.4 & 0.6$\pm$0.0 & 61.3$\pm$0.1 & 45.6$\pm$0.0 & 52.9$\pm$0.1 & 55.2$\pm$0.1 \\
        & Bi-GRU  & 98.8$\pm$0.0 & 93.8$\pm$0.1 & 55.5$\pm$0.1 & 16.6$\pm$0.2 & 0.7$\pm$0.0 & 64.1$\pm$0.1 & 47.8$\pm$0.1 & 55.8$\pm$0.1 & 58.9$\pm$0.1 \\
        & CAML & 98.8$\pm$0.0 & 90.7$\pm$0.3 & 58.6$\pm$0.1 & 19.3$\pm$0.1 & 0.6$\pm$0.0 & 66.3$\pm$0.1 & 50.3$\pm$0.0 & 58.5$\pm$0.1 & 62.4$\pm$0.1 \\
        & MultiResCNN  & 99.2$\pm$0.0 & 95.1$\pm$0.1 & 60.4$\pm$0.0 & 27.7$\pm$0.3 & 0.8$\pm$0.0 & 67.6$\pm$0.0 & 51.8$\pm$0.0 & 60.4$\pm$0.0 & 64.7$\pm$0.1 \\
        & LAAT  & 99.3$\pm$0.0 & 96.0$\pm$0.3 & 61.7$\pm$0.1 & 26.4$\pm$0.9 & 0.9$\pm$0.0 & 68.9$\pm$0.1 & 52.7$\pm$0.1 & 61.7$\pm$0.2 & 66.3$\pm$0.2 \\
        & PLM-ICD  & \bfseries 99.4$\pm$0.0 & \bfseries 97.2$\pm$0.2 & \bfseries 62.6$\pm$0.3 & \bfseries 29.8$\pm$1.0 & \bfseries 1.0$\pm$0.1 & \bfseries 70.0$\pm$0.2 & \bfseries 53.5$\pm$0.2 & \bfseries 62.7$\pm$0.3 & \bfseries 68.0$\pm$0.3 \\
        \hline
        \multirow{6}{*}{\shortstack[c]{MIMIC-IV \\ ICD-10}} & CNN & 97.5$\pm$0.1 & 87.9$\pm$0.4 & 47.2$\pm$0.6 & 8.0$\pm$0.4 & 0.3$\pm$0.0 & 60.3$\pm$0.1 & 45.7$\pm$0.1 & 47.3$\pm$0.2 & 48.2$\pm$0.2 \\
        & Bi-GRU  & 98.3$\pm$0.0 & 92.4$\pm$0.2 & 50.1$\pm$0.2 & 10.6$\pm$0.4 & 0.3$\pm$0.0 & 62.6$\pm$0.2 & 47.7$\pm$0.2 & 49.6$\pm$0.1 & 51.1$\pm$0.2 \\
        & CAML  & 98.5$\pm$0.0 & 91.1$\pm$0.1 & 55.4$\pm$0.2 & 16.0$\pm$0.3 & 0.3$\pm$0.0 & 66.8$\pm$0.2 & 52.2$\pm$0.1 & 54.5$\pm$0.2 & 57.4$\pm$0.2 \\
        & MultiResCNN  & 99.0$\pm$0.0 & 94.5$\pm$0.2 & 56.9$\pm$0.1 & \bfseries 21.1$\pm$0.2 & \bfseries 0.4$\pm$0.0 & 67.8$\pm$0.1 & 53.5$\pm$0.1 & 56.1$\pm$0.1 & 59.3$\pm$0.2 \\
        & LAAT  & 99.0$\pm$0.1 & 95.4$\pm$0.3 & 57.9$\pm$0.1 & 20.3$\pm$0.4 & \bfseries 0.4$\pm$0.0 & 68.9$\pm$0.1 & 54.3$\pm$0.1 & 57.2$\pm$0.1 & 60.6$\pm$0.2 \\
        & PLM-ICD   & \bfseries 99.2$\pm$0.0 & \bfseries 96.6$\pm$0.2 & \bfseries 58.5$\pm$0.7 & \bfseries 21.1$\pm$2.3 & \bfseries 0.4$\pm$0.0 & \bfseries 69.9$\pm$0.6 & \bfseries 55.0$\pm$0.6 & \bfseries 57.9$\pm$0.8 & \bfseries 61.9$\pm$0.9 \\
        
        \bottomrule
    \end{tabular}
\end{table*}

\subsection{Revised comparison}
The results of our revised comparison on MIMIC-III \textit{clean}, MIMIC-IV \textit{ICD-9}, and \textit{ICD-10} are shown in \cref{tab:fair_results}. Contrary to the originally reported results, Bi-GRU performs better than CNN in all metrics. Otherwise, the model performance ranking is unchanged from the original works. PLM-ICD outperformed the other models on all metrics and all datasets. The models previously reported as least performant improved the most. 

\begin{figure*}[t]
    \centering
    \begin{subfigure}[b]{0.44\textwidth}
        \centering
        \includegraphics[width=\linewidth]{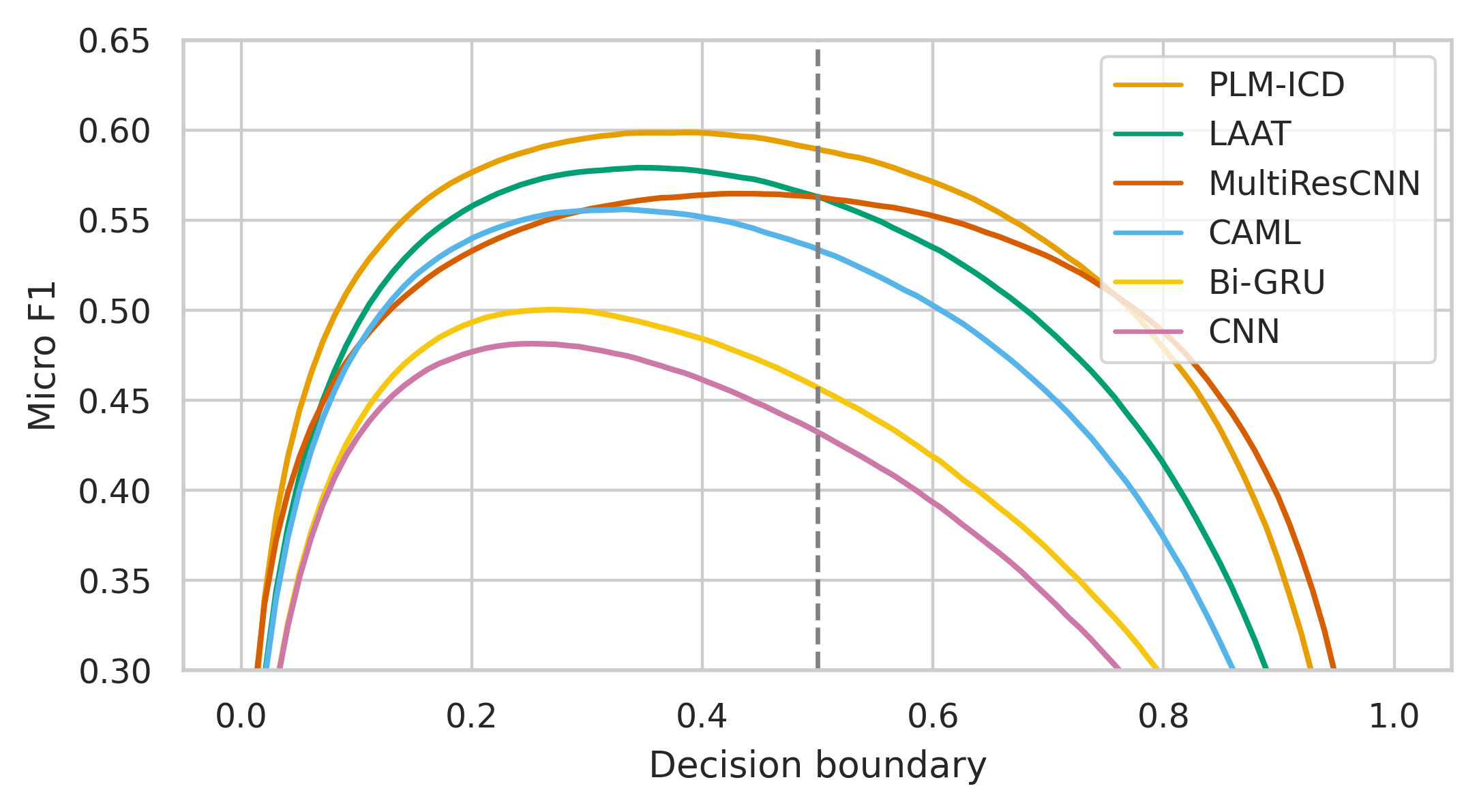}
    \end{subfigure}
    \hspace{10mm}
    \begin{subfigure}[b]{0.44\textwidth}
        \centering
        \includegraphics[width=\linewidth]{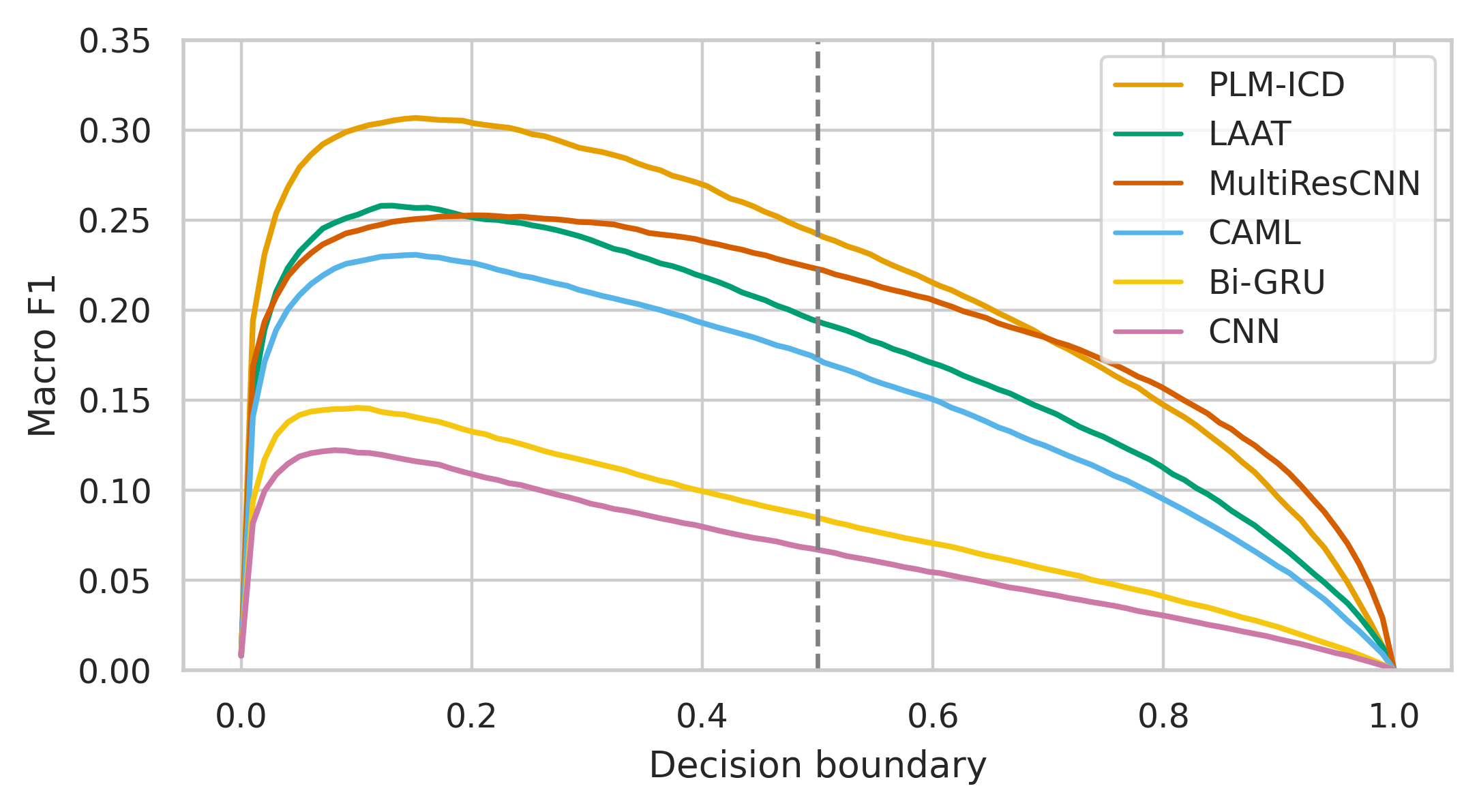}
    \end{subfigure}
    \caption{The relationship between chosen threshold and F1 score of every reproduced model in \cref{tab:reproduced_results}. The left figure shows the micro F1 score, and the right shows the macro F1 score. The models were evaluated on MIMIC-III \textit{clean}.}
    \label{fig:threshold_tuning}
\end{figure*}

The ablation study results are shown in \cref{tab:ablation study} for MIMIC-III \textit{clean}. Truncating the documents to 2,500 words instead of 4,000 had little impact on the performance. Using the hyperparameters from the original works degraded the performance substantially for CAML, Bi-GRU, and CNN but had a smaller effect on the other models. Not tuning the decision boundary had the largest negative effect on all models except MultiResCNN. In \cref{fig:threshold_tuning}, we plot the relationship between the decision boundary and F1 scores. LAAT and MultiResCNN perform similarly when using a decision boundary of 0.5. However, when tuning the decision boundary, LAAT outperforms MultiResCNN considerably.
Similar results were obtained on the other datasets. 

\begin{table*}[t]
    \centering
    \caption{Ablation study on MIMIC-III \textit{clean}. The numbers are the micro/macro F1 scores on the test set.}
    \label{tab:ablation study}
    \begin{tabular}{lcccccc}
        \toprule
         & PLM-ICD & LAAT & MultiResCNN & CAML & Bi-GRU & CNN \\
        \midrule
        Our result & 59.6/26.6 & 57.8/22.6 & 56.4/22.9 & 55.4/20.4 & 49.7/12.2 & 48.0/9.9 \\
        \hline
        Input length truncated at 2500 words & 59.4/26.2 & 57.6/22.3 & 56.0/23.2 & 54.8/19.7 & 49.4/12.0 & 47.9/9.8 \\
        No decision boundary tuning & 58.7/23.0 & 56.2/19.0 & 56.2/22.6 & 53.3/17.1 & 45.3/8.1 & 43.8/7.0 \\
        Original hyperparameters & 59.6/27.0 & 57.5/21.6 & 56.4/20.0 & 52.8/17.3 & 48.1/11.2 & 46.9/10.2 \\
        \bottomrule
        \end{tabular}
\end{table*}

\subsection{Error analysis}

\subsubsection{Amount of training data:}

\begin{figure*}[h]
    \centering
    \begin{subfigure}[b]{0.48\textwidth}
        \centering
        \includegraphics[width=\linewidth]{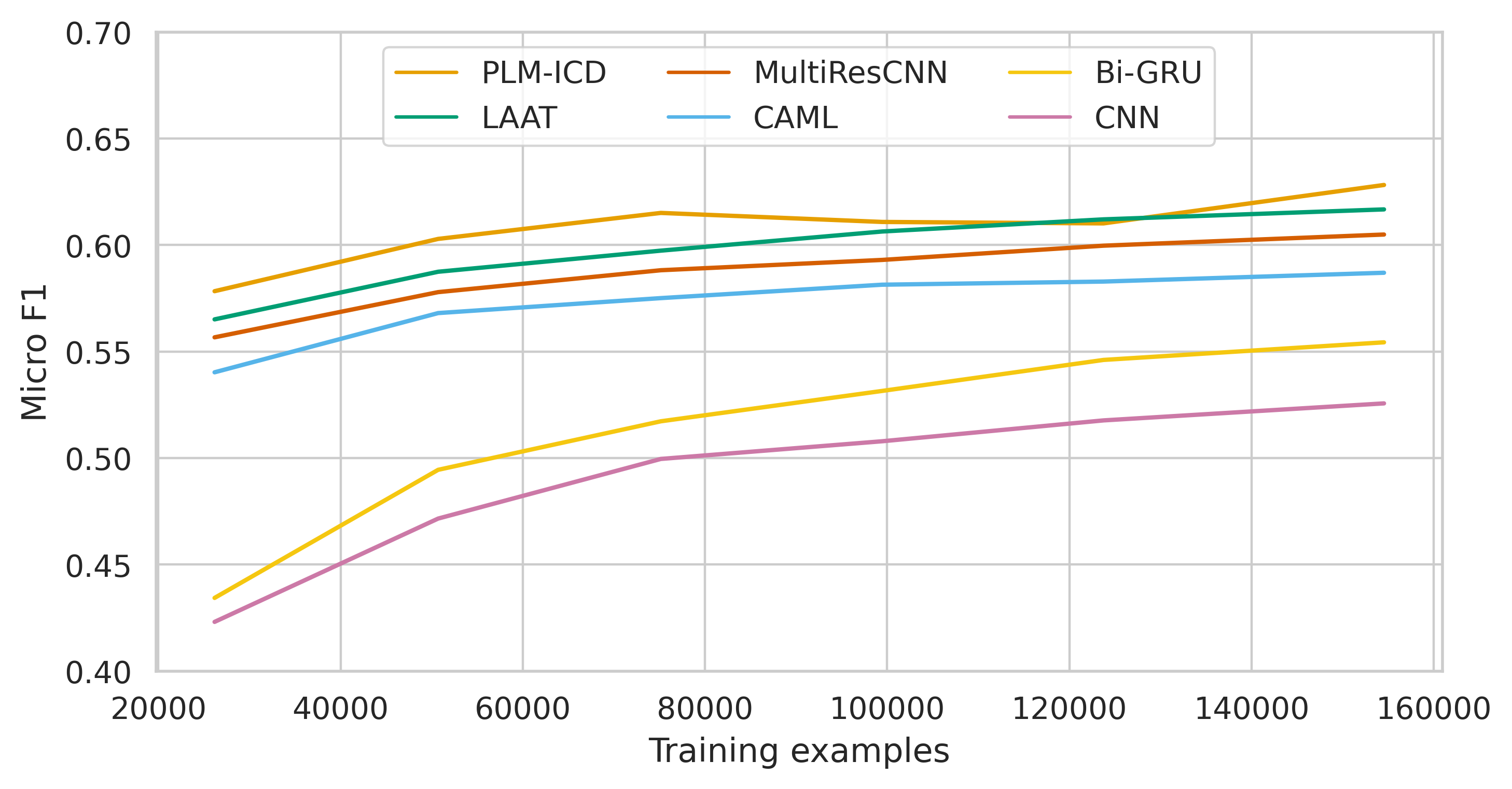}
    \end{subfigure}
    \hspace{3mm}
    \begin{subfigure}[b]{0.48\textwidth}
        \centering
        \includegraphics[width=\linewidth]{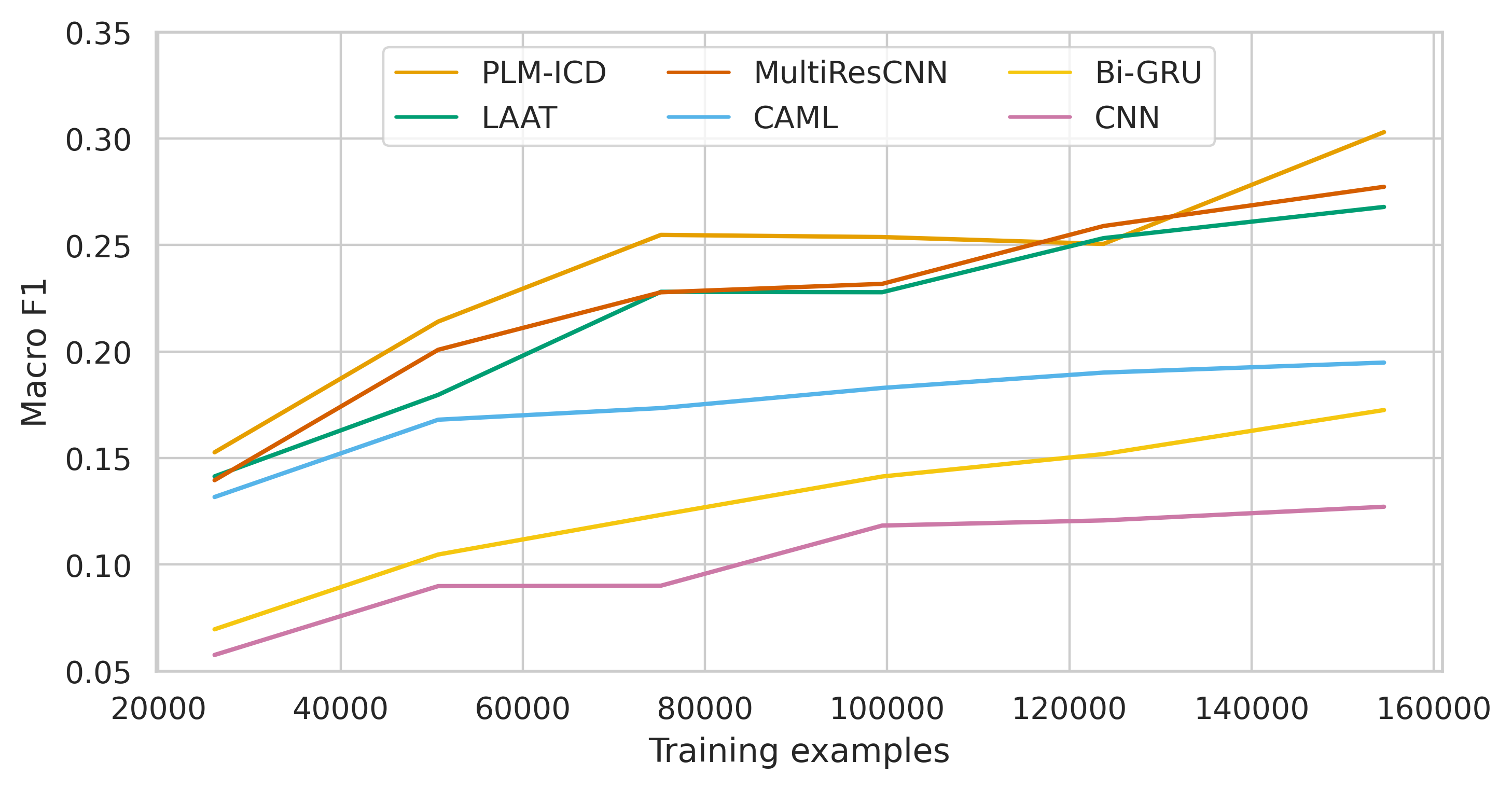}
    \end{subfigure}
    \caption{The relationship between the number of training examples and F1 score on MIMIC-IV \textit{ICD-9}. The left figure shows the F1 Micro score on the y-axis, while the right figure shows the F1 Macro score.}
    \label{fig:train_size}
\end{figure*}

\cref{fig:train_size} shows the relationship between the number of training examples and the micro and macro F1 scores for all models.
In most cases, increasing the training data had a larger effect on the macro F1 score than the micro F1 score, indicating more extensive improvements in rare codes than common codes. The curve for macro F1 is less smooth because the decision boundary was tuned on the micro F1 scores.


\subsubsection{Document length:}\label{subsubsec:document-length}

\begin{figure}[t]
    \centering
    \includegraphics[width=0.9\linewidth]{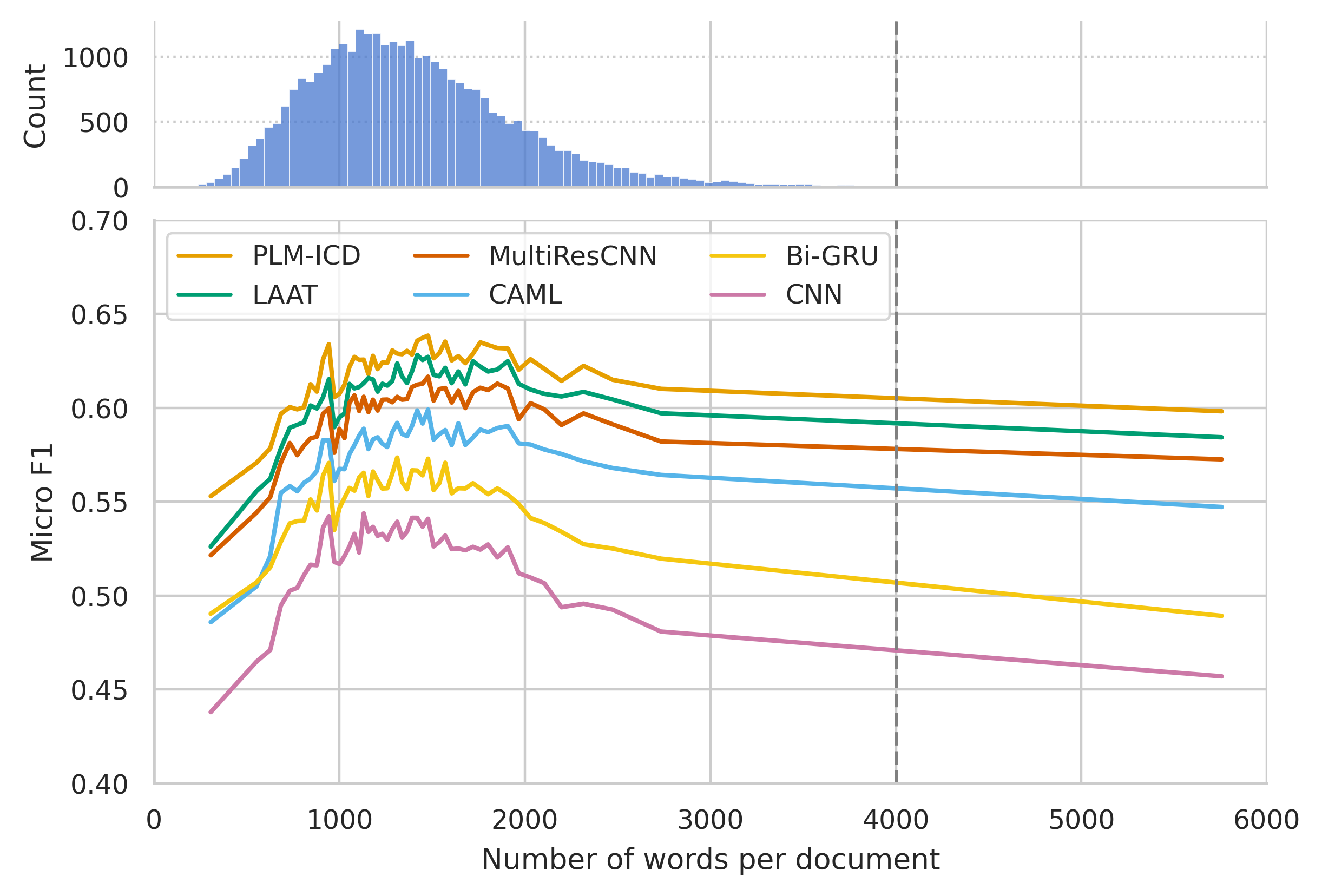}
    \caption{Relationship between the lengths of the clinical notes and the micro F1 score for each model on MIMIC-IV \textit{ICD-9}. The vertical line indicates the maximum length of the notes after truncation. The histogram at the top visualizes the document length distribution.}
    \label{fig:text_length}
\end{figure}

We plot the micro F1 score for all models as a function of the number of words per document in \cref{fig:text_length}. 
We note that all models underperformed on documents with fewer than 1000 words. 
By manual inspection, we found that most of these documents missed the information necessary to predict their labeled codes, leading to underperformance. 
In \cref{tab:correlations}, we list the Pearson and Spearman correlations. We excluded documents shorter than 1000 words to avoid confounding with missing information and longer than 4000 words due to the truncation limit. We observe a very small negative correlation between document length and micro F1 which matches the downward trend in micro F1, starting from approximately 1000 words in \cref{fig:text_length}. 
Although document length may itself be the cause of the slightly lower performance for long documents, there may be other factors correlated with document length impacting performance, such as the number of codes per document and code frequency.
As there are few long documents, the effect on average micro F1 for each dataset is negligible; hence, previous claims that long documents lead to poor performance in AMC could not be validated. Results on MIMIC-IV \textit{ICD-10} and MIMIC-III \textit{clean} were similar.


\subsubsection{Code analysis:}

\begin{figure}[t]
    \centering
    \includegraphics[width=0.9\linewidth]{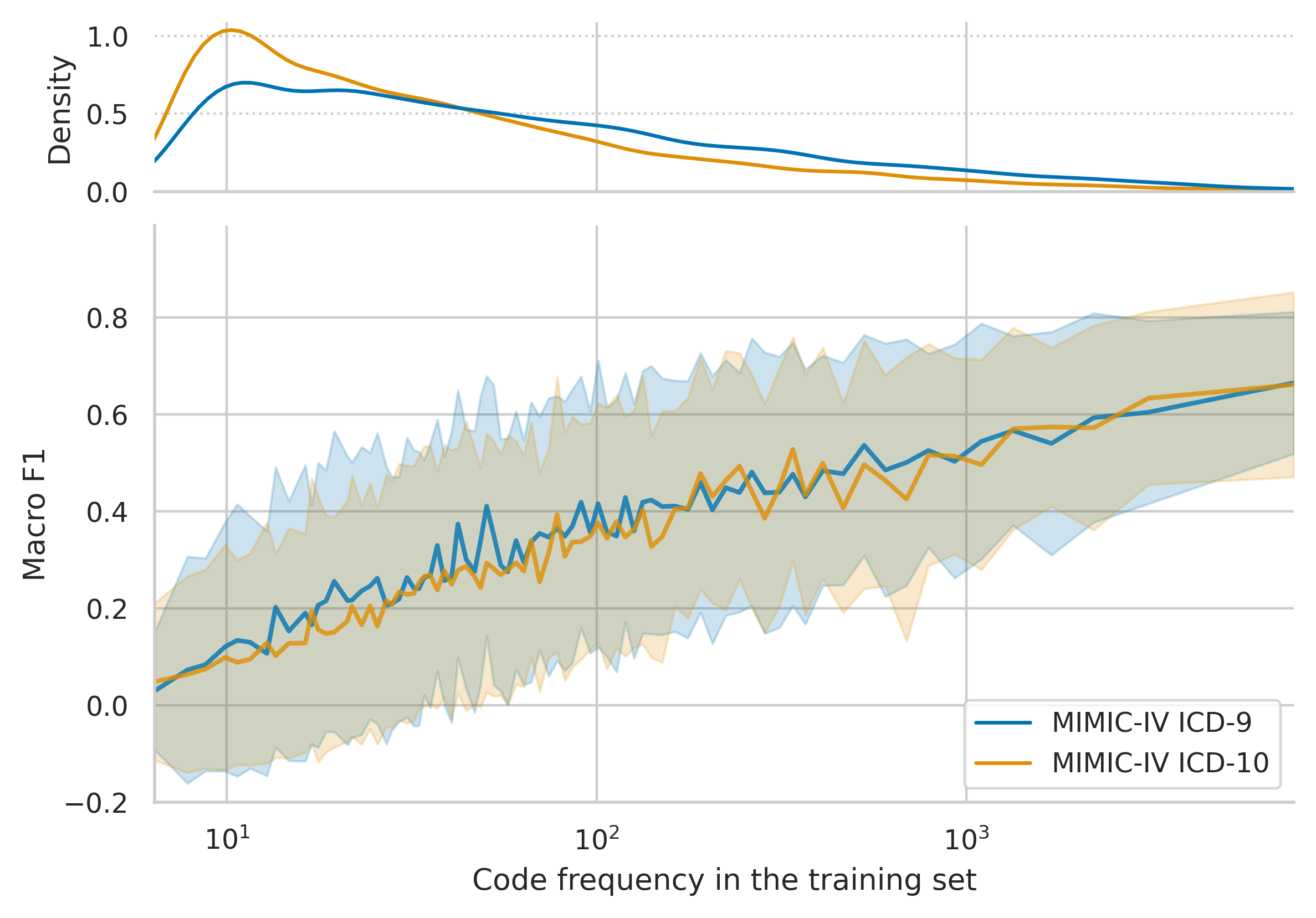}
    \caption{Relationship between the code frequencies in the training set and the macro F1 score for PLM-ICD on MIMIC-IV \textit{ICD-9} and \textit{ICD-10}. The shaded area indicates the standard deviation of the score computed for codes within the bin.}
    \label{fig:icd9_vs_icd10}
\end{figure}

\Cref{fig:icd9_vs_icd10} compares the best performing model, PLM-ICD, trained and evaluated on MIMIC-IV \textit{ICD-9} and \textit{ICD-10}. Similar results were obtained on MIMIC-III \textit{clean}.
The comparison shows the relationship between code frequencies in the training set and macro F1 scores. As shown in \Cref{tab:fair_results}, all models perform worse on \textit{ICD-10} compared to \textit{ICD-9}. However, \cref{fig:icd9_vs_icd10} demonstrates that performance on codes with similar frequencies is comparable between the two splits. This suggests that the performance differences in \cref{tab:fair_results} are due to \textit{ICD-10} containing a higher fraction of rare codes as shown in \cref{fig:mimiciv_code_dist,fig:icd9_vs_icd10}.

The Pearson and Spearman correlations between the logarithm of code frequency and F1 score are shown in \cref{tab:correlations} for MIMIC-IV \textit{ICD-9}. Similar correlations were observed for the other datasets. All the models show moderately high correlation confirming that performance on rare codes is generally lower than on common codes. To further our understanding of the problem, we computed the percentage of unique codes in each dataset that the models never predicted. As seen in \Cref{tab:missed_classes}, no model correctly predicted more than 50\% of the ICD-10 codes.





\begin{figure}[t]
    \centering
    \includegraphics[width=0.9\linewidth]{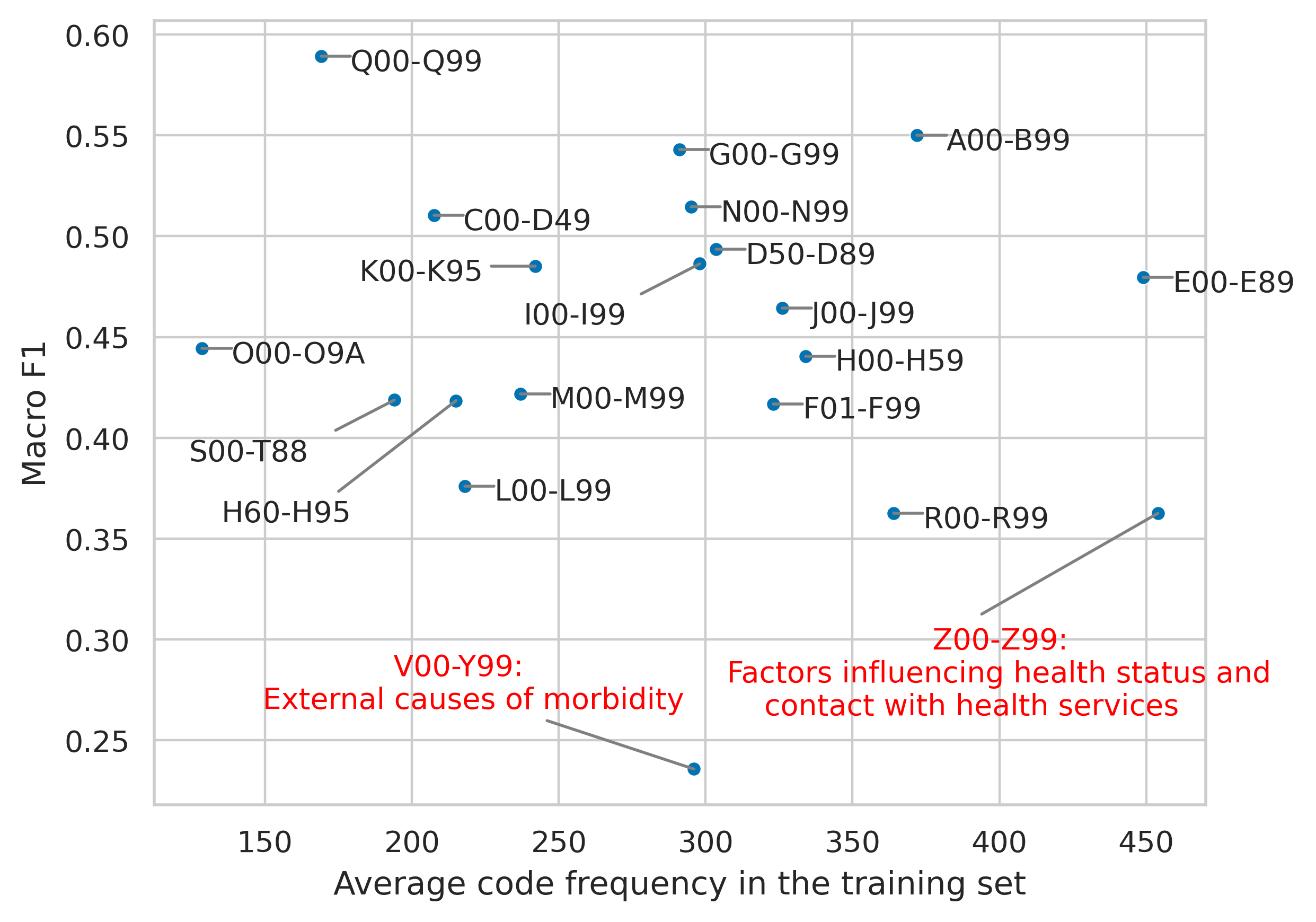}
    \caption{Performance of PLM-ICD on ICD-10 chapters. Only codes with more than a hundred occurrences in the MIIMC-IV ICD-10 training set were considered, leaving 20 chapters. We found Z00-Z99 and V00-Y99 to be the most challenging.}
    \label{fig:chapter_performance}
\end{figure}

\Cref{fig:chapter_performance} shows the performance of PLM-ICD on each ICD-10 chapter---the top-most level in the tree-like hierarchy. 
For our analysis, we limited the scope to only focus on the diagnosis codes. We also excluded codes with fewer than one hundred training examples to control for some chapters having many rare codes. 

Overall, PLM-ICD never correctly predicted 2,928 of the 5,794 ICD-10 diagnosis codes in our split. Of these codes, only 110 had over a hundred training examples, and 58 belong to only two of the 20 chapters in MIMIC-IV \textit{ICD-10}. Specifically, 45 belong to the chapter relating to ``external causes of morbidity'' (Z00-Z99), while 13 relate to ``factors influencing health status and contact with health services'' (V00-Y99). To further investigate why most non-predicted codes with more than 100 training examples belong to only two chapters, we manually inspected a selection of codes in these chapters, as described in the following.

The Z68 category, part of the Z00-Z99 chapter, contains codes related to the patient's body mass index (BMI). Codes within this category occur more than 17,000 times in the MIMIC-IV training data, but PLM-ICD never predicts 20 out of the 26 codes of Z68. One possible hypothesis is that PLM-ICD struggles with extracting the BMI from the discharge summaries, as all digits have been removed in the pre-processing. We found several other codes containing digits in the code descriptions that the model failed to detect, e.g., ``Blood alcohol level of less than 20 mg/100 ml'' (Y90.0), ``34 weeks gestation of pregnancy'' (Z3A.34), and ``NIHSS score 15' (R29.715). These observations support our hypothesis that removing digits in the pre-processing makes certain codes challenging to predict.

The Y92 category, part of the V00-Y99 chapter, contains codes related to the physical location of occurrence of the external cause. It is a large category of 246 unique codes occurring 27,870 times in the training set. The category is challenging due to locations being very specific. For instance, there are unique codes for whether an incident occurred on a tennis court, squash court, art gallery, or museum. We hypothesize that the level of detail in the discharge summaries does not always match the fine-grained code differences.

There are ten different codes in MIMIC-IV \textit{ICD-10} relating to nicotine dependence and tobacco use. The three most common are Z87.891 (``Personal history of nicotine dependence"), F17.210 (``Nicotine dependence, cigarettes, uncomplicated"), and Z72.0 (``Tobacco use"), with 26,427, 8,486, and 1,914 training examples, respectively. Among these, Z72.0 was the third most common single code in the training set that PLM-ICD never predicted correctly. PLM-ICD achieved an F1 score of 53\% for Z87.891, 51\% for F17.210, and 0\% for Z72.0 and all other nicotine-related codes. These findings suggest that when there is a class imbalance among highly similar codes, PLM-ICD is strongly biased toward the most frequent ones.

\begin{table}[t]
    \centering
    \caption{Correlation between the F1 score and the logarithm of code frequency and document length on MIMIC-IV \textit{ICD-9}. 
    As discussed in \cref{subsubsec:document-length}, we only considered document lengths between 1000 and 4000 words. All correlations are statistically significant ($p<0.001$).
    }
    \label{tab:correlations}
    \begin{tabular}{lcccc}
    \toprule
     & \multicolumn{2}{c}{Code frequency} & \multicolumn{2}{c}{Document lengths}\\
     \cmidrule(lr){2-3}\cmidrule(lr){4-5}
     & Pearson & Spearman & Pearson & Spearman \\
    \midrule
    CNN  & 0.61 & 0.68 & -0.09 & -0.08 \\
    Bi-GRU & 0.57 & 0.65 & -0.08 & -0.07 \\
    CAML & 0.56 & 0.60 & -0.03 & -0.03   \\
    MultiResCNN & 0.47 & 0.53 & -0.02 & -0.03  \\
    LAAT & 0.52 & 0.57 & -0.02 & -0.02 \\
    PLM-ICD & 0.48 & 0.52 & -0.02 & -0.02   \\
    \bottomrule
    \end{tabular}
\end{table}

 \begin{table}[h]
    \centering
    \caption{Percentage of ICD diagnosis codes in the test set that the models never predicted correctly.}
    \label{tab:missed_classes}
    \begin{tabular}{lccc}
    \toprule
    & MIMIC-III & \multicolumn{2}{c}{MIMIC-IV}\\
    \cmidrule(lr){2-2}\cmidrule(lr){3-4}
     & \textit{clean} &  \textit{ICD-9} & \textit{ICD-10} \\
    \midrule
    CNN  & 68.2 & 61.5 & 72.0\\
    Bi-GRU  & 65.0 & 54.3 & 67.1\\
    CAML & 52.8 & 57.0 & 62.0  \\
    MultiResCNN  & 48.8 & 40.3 & 53.5 \\
    LAAT  & 50.4 & 43.6 & 55.0\\
    PLM-ICD   & 44.3 & 39.3 & 51.8\\
    \bottomrule
    \end{tabular}
\end{table}

\section{Discussion}

\subsection{Lessons learned}
We found reproducing the results of CNN, Bi-GRU, CAML, and LAAT challenging. While we expected discrepancies due to random weight initializations and data shuffling, the differences from the original works exceeded our presuppositions. Our reproduced results were better than originally reported for Bi-GRU and CNN and worse for CAML and LAAT on most metrics. There have been multiple reports of issues in reproducing the results of \citet{mullenbachExplainablePredictionMedical2018}.\footnote{\url{https://github.com/jamesmullenbach/caml-mimic}} Additionally, most previous works did not report which version of MIMIC-III they used, and the code and hyperparameter configurations were not documented in detail. Therefore, we hypothesize that our results differ because previous works report incorrect hyperparameters or use an earlier version of MIMIC-III.


We showed that models previously reported as low-performing underperformed partly due to a poor selection of hyperparameters and not tuning the decision boundary. In our revised comparison, we demonstrated that training the models using our setup decreased the difference between the best and worst micro F1 scores by $5.8$ percentage points. \citet{mullenbachExplainablePredictionMedical2018} concluded that CNN outperformed Bi-GRU. However, in our revised comparison, Bi-GRU outperformed CNN on all metrics on MIMIC-III \textit{clean}, MIMIC-IV \textit{ICD-9}, and MIMIC-IV \textit{ICD-10}.

Even though MultiResCNN contains more parameters than CAML, \citet{liICDCodingClinical2020} concluded that MultiResCNN was faster to train because it converged in fewer epochs. However, this was only true when using the original setup where CAML converged after 84 epochs. We found that when using a learning rate schedule and appropriate hyperparameters, it was possible to train all the models in 20 epochs without sacrificing performance. Therefore, with our setup, CAML was faster to train than MultiResCNN.

We demonstrated that the macro F1 score had been underestimated in prior works due to the poorly sampled MIMIC-III \textit{full} split and the practice of setting the F1 score of all codes absent in the test set to 0. Since 54\% of the codes in MIMIC-III \textit{full} are missing in the test set, the maximum possible macro F1 score is 46\%. The previously highest reported macro F1 score on MIMIC-III \textit{full} is 12.7\% for PLM-ICD~\cite{kimReadAttendCode2021}. Using our corrected macro F1 score on the same split, PLM-ICD achieved a macro F1 score of 22.8\%. This large difference from previous state-of-the-art seems to indicate that all previous work on AMC used the sub-optimally calculated macro F1 score, including works not reproduced in this paper. Many studies use the macro F1 score to evaluate the ability of their models to predict rare codes \cite{kimReadAttendCode2021, yuanCodeSynonymsMatter2022}. If it has indeed been incorrectly calculated in these studies, some conclusions drawn in previous work regarding rare code prediction may have been misguided.

Multiple studies mention lack of training data, rare codes, and long documents as the main challenges of AMC \cite{dongExplainableAutomatedCoding2021,feuchtDescriptionbasedLabelAttention2021,huangPLMICDAutomaticICD2022,jiDoesMagicBERT2021,liICDCodingClinical2020,liuEffectiveConvolutionalAttention2021,moonsComparisonDeepLearning2020,pascualBERTbasedAutomaticICD2021,tengReviewDeepNeural2022,tengExplainablePredictionMedical2020, vuLabelAttentionModel2020, venkateshAutomatingOverburdenedClinical2023}. In the error analysis, we aimed to validate or falsify these assumptions. We found that rare codes were challenging for all models and observed that more than half of all ICD-10 codes were never predicted correctly. Furthermore, in \cref{fig:train_size}, we showed that when adding more training data, most models see a greater performance improvement on rare codes than on common codes. 
These findings suggest that medical coding is fundamentally challenged by a lack of training data that, in turn, gives rise to many rare codes.
We found that document length and model performance only exhibited a weak correlation. Specifically, the low number of very long documents was insufficient to affect the average performance on the dataset. 

\subsection{Future work}
We recommend future work within AMC use our revised comparison method, including stratified sampled splits of MIMIC datasets, corrected evaluation metrics, hyperparameter search, and decision boundary tuning to avoid reporting suboptimal or biased results. 
Furthermore, for AMC to become a viable solution for ICD-10, future research should focus on improving performance on rare codes while, in the shorter term, developing methods to detect codes that are too challenging for automated coding and, therefore, should be coded manually. Finally, while PLM-ICD outperforms the other models in this paper, the improvements are limited compared to the effect of pre-training in other domains \cite{mohamedSelfSupervisedSpeechRepresentation2022, linPretrainedTransformersText2021, baevskiWav2vecFrameworkSelfSupervised2020, devlinBERTPretrainingDeep2019, dosovitskiyImageWorth16x162020}.
Notably, there have been several unsuccessful attempts at using pre-trained transformers for medical coding \cite{jiDoesMagicBERT2021,gaoLimitationsTransformersClinical2021,michalopoulosICDBigBirdContextualEmbedding2022,pascualBERTbasedAutomaticICD2021,zhangBERTXMLLargeScale2020}. In future work, we want to investigate why pre-trained transformers underperform in medical coding.

\subsection{Limitations}
We presented findings and analyses on MIMIC-III and MIMIC-IV. It is unclear how our findings generalize to medical coding in real-world settings.
For instance, since MIMIC-III and IV contain data from the emergency department and ICU of a single hospital, the findings in this paper may not generalize to other departments or hospitals. 
For instance, discharge summaries from outpatient care are often easier to code than summaries from inpatient care as they are shorter with fewer codes per document \cite{zhangBERTXMLLargeScale2020, liuEffectiveConvolutionalAttention2021, tsengAdministrativeCostsAssociated2018}.

The medical code labeling of MIMIC is used as a gold standard in this paper. However, medical coding is error-prone, and, in many cases, deciding between certain codes can be a subjective matter \cite{nouraeiAuditNatureImpact2013, lloydPhysicianCodingErrors1985}. \citet{burnsSystematicReviewDischarge2012} systematically reviewed studies assessing the accuracy of human medical coders and found an overall median accuracy of 83.2\% (IQR: 67.3–92.1\%). \citet{searleExperimentalEvaluationDevelopment2020} investigated the quality of the human annotations in MIMIC-III and concluded that 35\% of the common codes were under-coded. Such errors and subjectivity in manual medical coding make model training and evaluation challenging and suggests that additional evaluation methods using, e.g., a human-in-the-loop, could be useful to increase the reliability of results. 

\section{Conclusion}
In this paper, we first reproduced the results of selected state-of-the-art models focusing on unimodal models with publically available source code. 
We found that model evaluation in original works was biased by an inappropriate formulation of the macro F1 score and treatment of missing classes in the test set. By fixing the macro F1 computation, we approximately doubled the macro F1 of the reproduced models on MIMIC-III \textit{full}. 
We introduced a new \textit{clean} split for MIMIC-III that contains all classes in the test set and performed a revised comparison of all models under the same training, evaluation, and experimental setup, including hyperparameter and decision boundary tuning. We observed a significant performance improvement for all models, with those previously reported as low-performing improving the most. 
We reported the first results of current state-of-the-art models on the newly released MIMIC-IV dataset \cite{johnsonMIMICIVFreelyAccessible2023, goldbergerPhysioBankPhysioToolkitPhysioNet2000} and provided splits for the \textit{ICD-9} and \textit{ICD-10} coded subsets using the same method as for MIMIC-III \textit{clean}.
Through error analysis, we provided empirical evidence for multiple model weaknesses. Specifically, models underperform severely on rare codes 
and, in contrast to previous claims, long documents only have a negligible negative performance impact. 
We release our source code, model parameters, and the new MIMIC-III \textit{clean} and MIMIC-IV \textit{ICD-9} and \textit{ICD-10} splits.$^\text{1}$

\begin{acks}
 This research was partially funded by the Innovation Fund Denmark via the Industrial Ph.D. Program (grant no. 2050-00040B, 0153-00167B, 2051-00015B) and Academy of Finland (grant no. 322653).
 We thank Sotiris Lamprinidis for implementing our stratification algorithm and data preprocessing helper functions.
\end{acks}

\bibliographystyle{ACM-Reference-Format}
\bibliography{references}

\appendix

\end{document}